\documentclass[submission,copyright,creativecommons]{eptcs}
\providecommand{\event}{SOS 2007} 

\usepackage{iftex}

\ifpdf
  \usepackage{underscore}         
  \usepackage[T1]{fontenc}        
\usepackage{latexsym}
\usepackage{tikz}
\usepackage{changepage}
\usepackage{makecell}
\usepackage{tcolorbox}
\usepackage{longtable}

\usepackage{graphicx}
\else
  \usepackage{breakurl}           
\fi

\title{Improving the Diproche CNL through Autoformalization via Large Language Models}
\author{Merlin Carl
\institute{EUF\\ Flensburg, Germany}
\institute{Institut f\"ur Mathematik\\
Europa-Universit\"at Flensburg\\
Flensburg, Germany}
\email{merlin.carl@uni-flensburg.de}
}
\def\titlerunning{Improving Diproche with LLMs}
\def\authorrunning{M. Carl}
\begin{document}
\maketitle

\begin{abstract}
The Diproche system is an automated proof checker for texts written in a controlled fragment of German, designed for didactical applications in classes introducing students to proofs for the first time. 
The first version of the system used a controlled natural language for which a Prolog formalization routine was written. In this paper, we explore the possibility of prompting large language models for autoformalization in the context of Diproche, with encouraging first results.

\end{abstract}

\section{Introduction}

The Diproche (``Didactical Proof Checking'')\footnote{Both the system's architecture and its name are inspired by the Naproche system, of which it is a kind of didactical ``offspring'', see \url{https://naproche-net.github.io/0}.} system is an automated proof checker for proofs in a controlled natural language (CNL) specifically adapted to elementary proving exercises in ``introduction to proof'' classes in first-year university education. Users can enter a proof in a controlled fragment of German and receive immediate feedback on logical correctness, type correctness, fulfillment of proof goals etc.; see \cite{C}, \cite{CK} for an introduction to the system and \cite{CLS} for a report on the experiences with the first version of the system. 

In spite of the impressive progress recently made with machine translation and the possibility of using this for the task of autoformalization (see, e.g., \cite{WBKU}), the Diproche CNL was implemented using classical techniques from computational linguistics, more specifically via a definite clause grammar written in Prolog. This was used to convert the natural language input into an internal list format, from which the proof obligations at each proof step are generated, to be passed on to an automated theorem provers (ATPs) in which those steps that should be accepted in the context of a specific exercise are hard-coded. The reason for this choice was to make the system as transparent as possible to the user: There should be no surprises concerning how the program interprets certain formulations, or which formulations it accepts. However, in practice, it quickly became apparent that this goal is in conflict with the other, similarly important, goal of providing a convenient CNL sufficiently rich for expressing proof texts in a way that resembles natural mathematical texts well enough to spare users the burden of learning a formalism on top of the difficulties they face when learning how to prove. Since the intended users are mathematical beginners with little to no experience with formal deduction, formal languages or proof calculi, it turned out that transparency is lost rather quickly. This leaves little reason to not take advantage of the possibilities of natural language processing based on machine learning techniques, such as pretrained language models. An obvious reason in favor of this approach is that developing, refining and changing such a prompt-based CNL is much easier and quicker than (re-)writing a formal grammar. In particular, making the model work in other natural languages is as easy as translating the natural language sentences in the prompt, no matter how (grammatically) different the new language is from the current one.\footnote{Not even this may be necessary: Prompted entirely with German sentences, our model was able to correctly process several (simple) sentences written in English, French and Chinese.}

In this paper, we begin to explore the possibilities of using the language model DaVinci-3 developed by OpenAI\footnote{See \url{https://openai.com/product}.} for various tasks in didactical systems that aim at teaching how to prove, in particular transforming natural language input into a formal representation that can be further processed by formal proof checkers. For this purpose, we built a prototype in which a Python-based preprocessing routine using DaVinci-3 was combined with the logical components of Diproche written in Prolog. We also consider a very recent version of GPT-4, which shows an even stronger autoformalization performance.

At first, the experiences reported in Avigad et al. \cite{APSARA} with using large language models for autoformalization appear to be discouraging for this plan: Only about 11 percent of the natural language inputs were formalized correctly (\cite{APSARA}, p. 3). Much better results were reported in \cite{WJLRSJS}, where more than 25 percent of the natural language inputs (which were problems for math competitions) were translated correctly into Isabelle (ibid., p. $1$). Still, for reliably checking even a simple natural language argument consisting of typically more than $10$ sentences with sufficient reliability to be of didactical use to beginner's students, anything considerably below a hundred percent is not good enough.

However, one needs to keep in mind that the approach in the works just mentioned is explicitly not to use a CNL, but to formalize sentences from everyday mathematical discourse. 
In contrast, our aim is by far more modest: Retaining the restriction to a small fragment of natural mathematical language, we want to use language models to (i) simplify the process of designing and converting to such languages and (ii) allow for more freedom in the specific choice of formulations compared with a CNL given by a formal syntax. The classical approach to translating between a CNL and a formal representation would be to implement a formal grammar for it, which is a quite cumbersome task. In contrast, our experience shows that, with certain qualifications, the Diproche CNL could be learned and improved with merely $71$ lines of example formalization by text-davinci-003. Even more impressingly, if one merely requires translation to a more standard first-order format that is easily convertible to the internal representation format used by Diproche, the same effect could be achieved for GPT-4-Turbo with a prompt merely containing the relevant to notation, but no examples.\footnote{Some qualifications apply here; see below.}

The new system architecture, adapted from the one given in \cite{C}, is as follows, where Python components are marked with ``Py'' and Prolog components are marked with ``Pr'', while ``LLM'' denotes the large language model used for the autoformalization:\footnote{An experimental system has been implemented using text-davinci-003. The integration of GPT-4-Turbo is currently being developed.}

\begin{figure}
\centering

\begin{tikzpicture}[scale=0.7]

\draw (10,3)--(18,3)--(18,4)--(10,4)--(10,3); 
\node at (14,3.5) {Interface (Py)};
\draw (12.5,3)--(12.5,4); 
\node at (11.5,3.5) {Input};
\draw (15.5,3)--(15.5,4); 
\node at (16.5,3.5) {Output};

\draw [->] (12,3)--(12,1); 
\draw [<-] (16,3)--(16,1); 


\draw (19,-2)--(25,-2)--(25,-1)--(19,-1)--(19,-2); 
\node at (22,-1.5) {LLM};
\draw [<->] (19,-1.5)--(17,-1.5); 

\draw (11,0)--(17,0)--(17,1)--(11,1)--(11,0); 
\node at (14,0.5) {Preprocessing (Py)};

\draw [<-] (14,-1)--(14,0); 

\draw (11,-2)--(17,-2)--(17,-1)--(11,-1)--(11,-2); 
\node at (14,-1.5) {Annotation (Py)};

\draw (11,-3)--(17,-3)--(17,-4)--(11,-4)--(11,-3); 
\node at (14,-3.5) {Post-Processing (Py)};

\draw [->] (14,-2)--(14,-3); 


\draw (11,-5)--(17,-5)--(17,-6)--(11,-6)--(11,-5); 
\node at (14,-5.5) {Text structure (Pr)};

\draw [->] (14,-4)--(14,-5); 

\draw [->] (17,-3.5)--(17.5,-3.5)--(17.5,-7.5)--(17,-7.5); 
\draw [->] (17,-3.5)--(20,-3.5)--(20,-9);  
\draw [->] (17,-5.5)--(19,-5.5)--(19,-9); 

\draw (11,-7)--(17,-7)--(17,-8)--(11,-8)--(11,-7); 
\node at (14,-7.5) {Generating ATP-Tasks (Pr)};

\draw [->] (14,-6)--(14,-7); 

\draw (11,-9)--(17,-9)--(17,-10)--(11,-10)--(11,-9); 
\node at (14,-9.5) {ATP (Pr)};

\draw [->] (14,-8)--(14,-9); 
\draw [->] (14,-10)--(14,-13); 

\draw (19,-9)--(25,-9)--(25,-10)--(19,-10)--(19,-9); 
\node at (22,-9.5) {Goal Check (Pr)};

\draw [->] (20,-10)--(20,-13); 

\draw (13,-13)--(21,-13)--(21,-14)--(13,-14)--(13,-13); 
\node at (17,-13.5) {Feedback (Pr)};

\draw [->] (21,-13.5)--(26,-13.5)--(26,3.5)--(18,3.5); 

\end{tikzpicture}
\caption{Flowchart of Diproche with integrated LLM}
\end{figure}
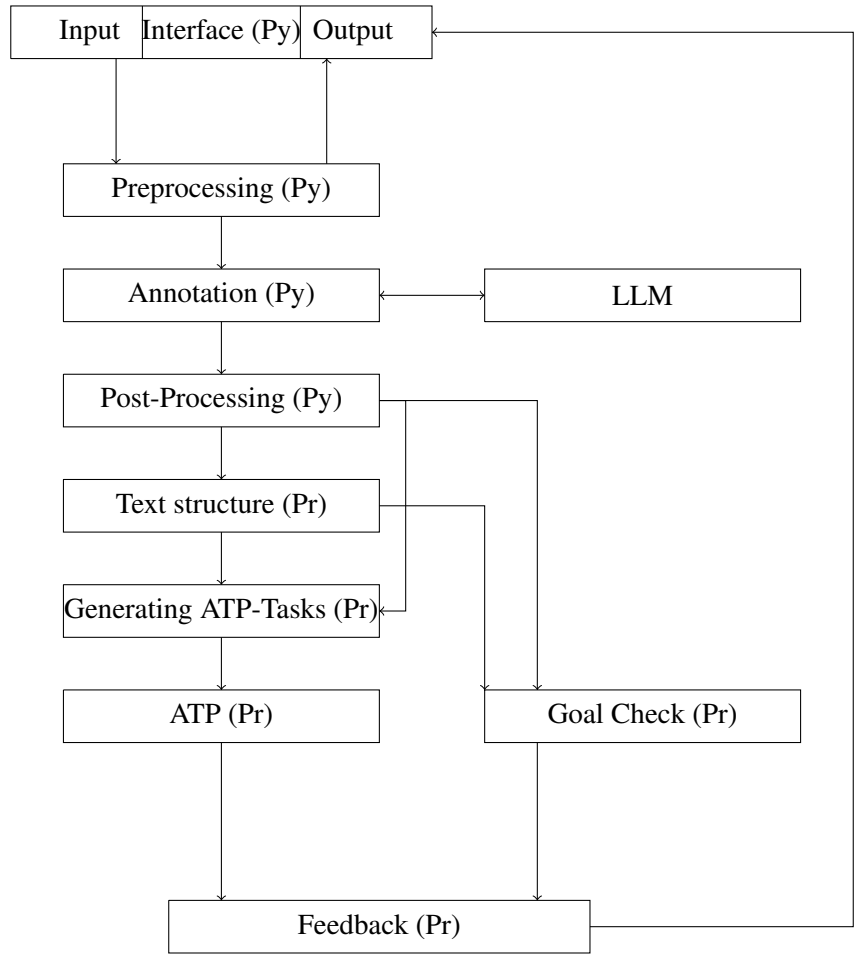

\section{Autoformalization and Proof-Checking}

A naive approach to using autoformalization in automated proof-checking is the following: Each natural language sentence corresponds to some formula in first-order logic. By translating each sentence separately, one obtains a formalization of the whole text, which can then be given to an automated theorem prover for verification. This view, however, ignores a great deal of well-known features of natural language mathematics:\footnote{Cf., e.g., \cite{CFKKSV}, p. 171; for a detailed treatments of the specifics of the language of mathematics, see Ganesalingam \cite{Ga}.}

\begin{enumerate}
\item Sentences have different functions, such as goal announcements, deductions, assumptions, annotations. For a (logical) check of the text, these need to be identified, along with the content. 
\item Sentences may have content that is not immediately expressible in first order logic. Consider ``Since $a=b$, it follows that $a^2=b^2$''. The claim here is apparently that, at this point of the argument, it is established that $a=b$ and that, from this, $a^2=b^2$ can be deduced. It would certainly be wrong to formalize this as $(a=b)\rightarrow(a^2=b^2)$, since then, only the implication would be established, while it is its conclusion that the sentence is claiming. However, omitting the ``Since $a=b$'' would considerably distort the sentence's meaning: We do not want a sentence like ``Since grass is green, it follows that $a^2=b^2$'' to be marked as correct. Another example would be a sentence containing multiple conclusions, such as ``Now we have $A$, so we get $A\vee B$, and thus also $C\rightarrow(A\vee B)$''. Thus, the formal representation of a sentence will need to use means beyond mere first-order logic. 
\item The whole text has a structure. It may contain subproofs, variables and assumptions are introduced for certain parts of the argument and gone in others.\footnote{See, e.g., Cramer \cite{Cr}, p. 255.} This overall structure needs to be taken into account when formally representing proofs. 
\item Elliptic sentences that gain their meaning from context: ``We show that there are infinitely many primes. Suppose otherwise.''; ``Thus, there is a line passing through $P$ and $Q$. Call it $l$.''; ``Hence, $n$ has at least one prime divisor. Pick one, and call it $p$.''; ``So we have $x\in A$. Consequently, it is even.''\footnote{Cf, e.g., \cite{SK}, p. 7-8.} 
\end{enumerate}

Thus, a naive ``sentence by sentence''-formalization is not enough as a basis for automated proof-checking. Along with a formalization of content, one needs to identify the function of the sentence, which is a task for automated text classification, one needs a formalism capable of representing figures such as justifications that do not appear in first-order logic (and the autoformalization routine needs to translate to this formalism), and one needs some kind of structural markers to identify the scope of declarations and assumptions, and, when formalizing a sentence, it must be possible to take into account earlier sentences as context.

\section{Prompting and Training Large Language Models}

The pre-trained language models offered by OpenAI can be adapted to a specific task in two different ways: Prompting and fine-tuning. While prompting can simply mean writing a description of the task at hand (such as ``Write a birthday card for the person named in the input''), in the case of autoformalization, it is best done by offering a carefully chosen series of examples.\footnote{This approach is also taken in Avigad et al. \cite{APSARA}.} Prompting leaves the model internally unchanged; intuitively, one could regard it as writing a (long) question. Fine-tuning, in contrast, means actually modifying (training) the language model with an appropriate data set. This option is currently only available for language models considerably weaker than text-davinci-003; in order to achieve satisfying result, a considerably larger amount of examples is required.\footnote{According to OpenAI \url{https://platform.openai.com/docs/guides/fine-tuning/preparing-your-dataset}
, one should ``aim for at least $\sim$100 examples per class'' for the classification task alone of associating sentences with their logical function, which would in our case amount to 700 sentences; the actual formalization task being by far more complex, one would likely need several thousand example sentences for reasonable results.} Since such data is currently unavailable for Diproche, and also somewhat cumbersome to generate, we will only consider prompting in this paper, which seems to offer a quick and easy way to achieve an impressive level of autoformalization sufficient for basic didactical applications.\footnote{A word of warning is in order here: A prompt length currently cannot go beyond 4000 tokens, which is quite limited.}

\subsection{The Diproche CNL and the Internal List Format}

The Diproche CNL is explained in some detail in \cite{C}. Here, we recall some basic features. The Diproche CNL is a fragment of natural mathematical German, comprising typical ways to express assumptions (``Suppose that $x$ is even''), claims (``Hence, $x+1$ is odd''), variable declarations (``Let $k$ be an integer''), goal announcements (``We will show that $x$ is a square'') and annotations (``Proof:'', ``qed'', ``Case $1$:'' etc.); further sub-types include declarations combined with assumptions, such as ``Let $k$ be an integer such that $n=2k$'' (which are existentially loaded and are in need of verification) and justified claims (``Since $x=3(a+b)$, $x$ is a multiple of $3$''). The sentences\footnote{In the relevant sense here, ``sentence'' includes annotations.} written in this CNL are converted into an internal list format whose crucial ingredients are a list of the of variables occuring in the sentence, its type (assumption, declaration, claim, annotation,...) and its actual content (which can be empty, as in the case of annotations). Thus, the sentence ``Therefore, $x$ is even'' would be translated as [[x],beh,[even,[x]]].

When processing formulations such as ``Suppose not'' or resolving anaphors, prior sentences need to be taken into account as the context in which a certain sentences is to be translated. A typical line of our prompt looks like this:\footnote{Translated into English for the convenience of the reader; the actual prompt lines are German.}

\begin{center}

context:\{We show that the intersection of $A$ and $B$ is a subset of the union of $A$ and $B$.\} Suppose not. \#   [[A,B],ang,[not,[[A,cap,B],subseteq,[A,cup,B]]]]\S

\end{center}

Here, the part ``context:\{\}'' contains the relevant context, the next part (``Suppose not'') is the sentence to be translated, \# serves to separate the natural language sentence from its formalization, then we have the formalization in the internal list format and finally \S\  as the stop symbol, which prevents the language model from generating further text, such as more examples of natural language sentences and their formalizations in the same spirit. The examples also included ungrammatical and formally invalid sentences, for which the translation was ``invalid'' (``ung\"ultig''). If the formalization routine leads to this result, the process is stopped and the sentence in question is reported to the user as not processable.

\subsection{First Experiences}

The first experiences with prompting DaVinci-3 for autoformalization in Diproche were encouraging: After only a few examples, the model had ``grasped'' the extraction of variables along with the classification and even offered (usually sensible) completions in the ``spirit'' of the given examples although no specific example for the case at hand was given; for example, after learning that ``$x$ is even'' was to be formalized as [even,[x]], it drew the obvious analogy for ``$x$ is odd'' or ``$x$ is prime''. It correctly dealt with formulations that it had not seen in the examples -- for example, after having seen that $x\in X$ was to be turned into [$x$,in,$X$], it correctly processed sentences like ``$x$ is an element of $X$'' or ``$X$ contains $x$'';\footnote{Although our actual prompts were written in German, we offer English translations here for the sake of the reader. Although it would surprise us if it was otherwise, we therefore cannot guarantee that the results can be reproduced with prompts written in English.} similarly, after being given one example of how to formalize an assumption, it correctly identified a variety of ways to formulate assumptions -- including some that, such as ``Gesetzt, es w\"are der Fall, dass'' (``Let it be the case that''), were intentionally chosen to be somewhat uncommon. It even had a considerable success rate when, after a series of German prompts, it was given a sentence in English, French, Italian or Chinese; a system once developed in this way can thus be easily made available in other languages as well. Even without prior examples, the model exhibited the ability introduce variable names not given in the text and pick them in a sensible way; thus, for example, ``Every natural number has a prime divisor'' was formalized as [all,[n,in,nat],[exists,p,[[prime,p],and,[divides,[p,n]]]]]; in particular, in no instance was a variable name used for different variables. In most cases, the model was able to resolve anaphors, taking advantage not merely of grammatical categories, but even of content: For example, for the input ``Hence $x$ is an element of $A$. Thus, it is even. Consequently, it cannot be empty.'', the first ``it'' was formalized as $x$, while the second was formalized as $A$. Turning formulas into the internal list format was also ``learned'' reliably along the way. Likewise, the model could correctly deal with elliptic formulations such as ``We will show that $A=B$. Suppose not.'' of ``Hence, there exists $k$ such that $n=2k$. Pick one.''. We also observed that a rather common input format for automated theorem provers, namely THF\footnote{See, e.g., \url{https://www.tptp.org/Seminars/THF/Contents.html} or \cite{BRS}.} was already ``known'' to DaVinci-3; when merely asked to ``formalize'' various statements without given specific examples, it generated THF formulas.\footnote{A similar ``surprise'' is reported in \cite{WJLRSJS}, p. 1.} This, however, is not of immediate relevance for our purposes, since Diproche uses its own internal format.

Still, there were issues, most of which, however, could be resolved to our satisfaction:

\begin{enumerate}
\item Along with the requested formalization, the model occasionally generated extra example pairs of natural language and formalization of its own. This could be prevented by introducing \S\ as a stop symbol and putting this after each formalization.
\item Even so, the model sometimes added to the natural language input rather than merely formalizing the given expression. This was resolved by separating the natural language expression from the formalization by an \# and making \# a part of the input for each request. 
\item When the input consisted of several sentences, the model frequently misrepresented later sentences, perhaps according to ``expectations'' of what these should have been rather than what was actually there. Only giving it one sentence per time was not an option, since this would have ruled out using the abilities of the model to refer to context, e.g., in the resolution of anaphors, or in processing such constructions as ``We will show that $p$ is prime. Suppose not.'' (see above). This was solved by explicitly labeling the preceding sentences as ``context''. The model was prompted with examples of sentences that could not be formalized without further context, in which case it should return the error message ``missing context''. In order to minimize such unwanted interference, the autoformalization routine works with the minimal amount of context that is required for the sentence at hand: Thus, given a list of sentences $[S_{1},S_{2},...,S_{n}]$, it will, in the $i$-step, attempt to formalize $S_{i}$ without invoking context (i.e., using the empty context). If this yields to a ``missing context'' error, $S_{i}$ is tried again, this time with context $\{S_{i-1}\}$. If the error persists, the earlier sentences are added one by one; when all earlier sentences up to $S_{1}$ were added without success, the formalization attempt is stopped unsuccessfully.
\item Even when the correct formalization was given in several examples, the model occasionally choose to express it differently, i.e., express $A\cup B$ as [A,union,B], rather than [A,cup,B]. Since these cases were of a limited and surveyable number, this problem could be dealt with by a postprocessing routine.
\item There were also difficulties to distinguish between implications (``If $x$ is even, then $x+1$ is odd'') and justifications (''Now $x+1$ is odd, because $x$ is even''): rewriting the examples and adding the tag ``justification'' improved the performance, but attained nothing near perfect accuracy. We therefore decided to drop phrases containing justifications from the CNL in the first version.
\end{enumerate}

Moreover, it soon became apparent that converting all types of formulas occuring in any of the sub-areas currently available in Diproche -- in particular, propositional logic, Boolean set theory, axiomatic geometry and elementary number theory -- was too much to ask from a model prompted with only 4000 tokens. We thus decided to further specify the task by writing separate prompts for each of these areas; when requested, the system would then use the model for the area to which the current proof text belongs. For the first experiments reported here, we concentrated on writing a prompt for Boolean set theory, an area for which several example Diproche texts are available and which also forms an important part of the beginner's lectures taught in Flensburg. 

Besides text-davinci-003, we also had the opportunity to test an ``assistant'' based on a recent version of GPT-4. An assistant is a chatbot whose behaviour can be controlled by a prompt describing its intended functioning.\footnote{See, e.g., \url{https://platform.openai.com/docs/assistants/how-it-works/agents}.} In order to keep the prompt length limited, we did not insist on a direct translation into the internal Diproche format, but were instead content with a standard first-order format that is easily automatically translatable into the required format. It turned out that, with a prompt explaining in detail the desired output format, the performance of the assistant was satisfying (see below) even without presenting a single example. However, it should be noted that the model's responses still depend on the previous course of the dialogue, so that answers to earlier requests play a role similar to the examples given to text-davinci-003. Thus, while the model was able to autoformalize the given statements with a rather high success rate, this might have been different had the statements been given in a different order. Thus, adding a set of examples representative of the task at hand seems advisable also for this option.

\section{Performance on Typical Diproche Texts}

To see whether the prompted language model would perform, we tested it against three solutions for set-theoretical exercises written in the Diproche CNL, and also modified versions of these solutions that contained mistakes. Not counting these variants, and only considering sentences that were actually passed on to the language model\footnote{Some standard annotations are processed separately and not given to the language model.}, these texts contained 33 sentences, all of which were processed correctly. Since these texts were typical texts that users of the system would be expected to write, this confirms that the prompted language model could serve as a replacement for the Diproche CNL in a practical setting. 

In order to evaluate the performance of the GPT-4-based assistant, we used $50$ example sentences from the area of Boolean set theory. The model was then asked to identify the type of the sentence (declaration, assumption, claim, declaration with additional assumption) and to provide a formalization. The results of this experiment can be found in a table in the appendix. In order to make it easier for the reader to evaluate the quality of the formalizations, we worked with English sentences, although the Diproche CNL is a fragment of German. As experiments suggest, the performance on German translations of the given sentences did not substantially deviate.  

We also asked several mathematicians to write up clearly structured solutions using short and simple sentences to several sample exercises, but without mentioning any particular syntax. Our goal is to evaluate how much of these solution texts will be processed correctly by the model in order to quantify the ``naturalness'' of the ``learned CNL''. The results of this will appear in future work.

These results show that, at least in the field of Boolean set theory, the ``learned CNL'', covers most of what the hard-coded Diproche-CNL for this purpose had to offer (we recall that ``justified claims'' were excluded in this investigation). On the other hand, it offers a much greater degree of freedom of expression; in particular, natural language variants of simple formal expressions, such as ``$x$ is an element of the intersection of $A$ and $B$'' are usually processed correctly, which would be somewhat cumbersome to obtain with a formal syntax.\footnote{We remark, though, that this did not always work reliably: For example, we observed one case where the model confused ``the intersection of the complements of $A$ and $B$'' with ``the complement of the intersection of $A$ and $B$''.} Moreover, it turned out to be rather tolerant with respect to minor misspellings or typos, in contrast to the Diproche CNL, the orthographical strictness of which had apparently been a source of frustration for several students. This is a clear advantage of using large language models over using hand-crafted formalization routines.

To get a clearer picture of the prospects of this approach, it would certainly be desirable to have a systematic statistical evaluation of the system. This, however, would in fact of limited value due to the following reason: Since the models used above are continually modified, the performance measured at one point of time can be vastly different from the performance a few months later, even when evaluated using the exactly same requests.\footnote{See, e.g., \cite{GPTChanges}, which indicates that the performance of GPT-4 and GPT-3.5 considerably \textit{dropped} from March to June 2023 in several areas.} We are planning such an evaluation after the system has been changed to work with local LLMs that can be kept stable over time (see also the next section). Moreover, since the preparation of the first version of this paper (in spring 2023) and the present version (in December 2024), new LLMs have become available that show a strongly improved performance for autoformalization tasks relevant to our purposes. 

To give the reader at least some impression of the possibilities, we discuss here briefly the 50 above-mentioned example sentences processed with an OpenAI API-assistant based on the (currently experimental) model GPT-4-Turbo.\footnote{Available under the name gpt-4-1106-preview.} The precise prompt and the table of results can be found in the appendix. Note that the prompt does not contain any examples. Of these $50$ examples, $49$ were processed correctly, leading to a success rate of 98 percent. The resolvation of anaphorical expressions worked well in most cases, see, e.g., (13), (19), (23), (24). The exception was example (30) ``From this, we get that, if $A$ is not empty, then $B$ is'', where the anaphorical expression ``then $B$ is'' was wrongly interpreted as referring to ``not empty'', while it would usually be understood as ``empty''. Phrases such as ``exactly one'' were correctly interpreted (46). The results also show that a variety of formulations was correctly processed, which considerably goes beyond the Diproche CNL, including in particular term description in natural language (13), (39), (41). 
Still, there are several points to be noted: First of all, the input format in this case did not take into account context. Due to this, sentences such as ``Let $x$ be an element of $X$'' (sentence 11) become ambiguous in their role, being either variable declarations (if $x$ appears for the first time in this sentence) or plain assumptions (if $x$ was introduced earlier on). Similarly, sentence $16$ (``Let $A$ be a subset of $B$'') was interpreted as a variable declaration of $A$ (but not of $B$). While this is indeed a plausible reading, there are of course contexts where it would be wrong. 
In order to fix this, some kind of context needs to be provided, at least in the form of declared variables available at a certain point in the text. 
While sentence $6$ was correctly formalized, the use of an existential quantifier for formalizing the phrase ``non-empty intersection'', rather than merely writing $G\cap H\neq\emptyset$, would in many contexts lead to difficulties in the further processing. 
In sentence (36), ``whenever'' could be read as an implicit universal quantifier, in which case the given formalization would be missing the quantifiers. However, since the formalization would be processed correctly in the context of Diproche, we regarded this as correct.

\section{Conclusion and Further Work}

The work reported in this paper indicates that prompting large language models such as DaVinci-3 is apparently a good ``quick and dirty'' way for developing and improving controlled natural languages for didactical systems such as Diproche which aim at automated proof verification at the beginner level; in particular, such languages tend to be more flexible and error-tolerant than those obtainable by classical methods with reasonable effort. Clearly, this merely scratches the surface of the possibilities that large language models open for such systems, and which will be the subject of further work. Moreover, the purpose of the present paper is merely to present the general concept and argue for its prospects; an accurate evaluation of the didactical advantages of using LLMs over formal grammars will be done once a system version suitable for the actual employment in teaching has been obtained.

On the one hand, the quality and reliability of autoformalization, which is currently mainly hindered by the bound on prompt length, could most likely be considerably improved by ``specialization'', i.e., writing prompts for various sub-tasks, classifying the input accordingly and then handing them to the appropriate sub-module. 

On the other hand, there is a number of other tasks relevant for such systems, which could conceivably be treated with large language models, including the following:

\begin{enumerate}
\item Immediate feedback on the proof text. Our experiments indicate that GPT is currently quite poor and unstable in evaluating the logical coherence of a proof text. However, for other kinds of feedback, such as stylistic remarks, it may be more useful.
\item Generating hints. While GPT is currently not able to reliably solve basic proving exercises, the texts it generates quite often tend to have the right overall structure. This could be used to provide hints for users how to approach a certain problem.
\item Mistake diagnosis. The current Diproche version includes an ``Anti-ATP'', a logical (and algebraic) mal-rule library that codifies typical false deduction steps (such as deducing $\neg{B}$ from $\neg{A}$ and $A\rightarrow B$) and reports them to the user; the hope is that this can help becoming aware of such mistakes and eventually to avoid them. Using large language models for mistake diagnosis at least for algebraic manipulations may lead to a more ``semantical'' approach to this task, in contrast to the current one based on formal patterns.\footnote{For example, both $(x+y)^{2}=x^{2}+y^{2}$ and $(x+1)^{3}=x^{3}+1$ are currently identified as an instance of a distributive use of exponentiation, but $(x+y+z)^{2}=x^{2}+y^{2}+z^{2}$ is not; in contrast, DaVinci-3 was able to recognize the third example as an instance of the same mistake, even though only examples with two summands were given to it in the prompt.}
\end{enumerate}

Concerning the practical use in (large) teaching situations, however, the following should be considered: 

\begin{itemize}
\item Since DaVinci-3 and similar large language models cannot be run locally, each checking requires a considerable amount of web traffic between the user's device and the servers on which the model is hosted. Compared with the good old-fashioned parsing approach, this takes noticably more time, in particular depending on the internet connection. In our experiments, the time from demanding feedback to receiving feedback typically went up from hardly noticable to at least 30 seconds.
\item The servers to which the formalization requests are sent are of limited capacity and, if too many requests are sent in a certain amount of time, will refuse them. This already happened to the author frequently during development. With a considerable amount of students frequently using the system, it can be expected to happen regularly.
\item Also, requests aren't free: One call to DaVinci-3 with the full amount of 4000 tokens costs about 8\textcent.\footnote{2\textcent per 1000 tokens, see \url{https://openai.com/pricing} (accessed 12.03.2023).} With a class of 250 students who are supposed to use the system regularly for their homework and make frequent intermediate checks when working on the exercises -- which is the whole purpose of the system -- this can quickly get expensive: Four such exercises with an average of five intermediate checkings would lead to a weekly price of 400\$.
\item Moreover, this approach has all the disadvantages of relying on external services: The underlying model may be changed, resulting in unexpected behaviour, or discontinued altogether; prices can go up; a certain amount of user data is sent to external servers, which may potentially lead to privacy issues etc. 
\item In particular, as \cite{GPTChanges} indicate, the continued modifications to GPT have led to a considerable \textit{decrease} in certain mathematical abilities in three months. Similarly, we observed that certain requests that consistently worked fine at some point of time did so no longer some months later. 
\item The ``learned CNL'', although quite accurate most of the time, is still somewhat unreliable and occasionally shows surprising behaviour. Thus, for example, the sentence ``Thus, $x$ is an element of $A$ or $x$ is an element of $A$.'' was reported by text-davinci-003 as ``invalid'', while it worked perfectly well when replacing one of the $A$'s by a different letter. This is likely due to the fact that such constructions are extremely uncommon in natural language. Moreover, there were instances where claims in the form conditional constructions were mis-identified as assumptions. In one case, a sentence was processed wrongly after changing the variable names. These difficulties can certainly be overcome by providing more examples, but it should still be kept in mind that, compared with hard-coded parsers, a certain amount of reliability is lost. 
\item A potential problem that may arise in use is that a ``less controlled CNL'' may make it more difficult for users to develop a feeling for what will be understood and what not. Whether this is an actual problem will have to be evaluated empirically. 

\end{itemize}

To sum up, experiences indicate that large language models can be fruitfully and easily applied in developing CNLs for proof checkers for didactical applications. However, for practical applications, the aim should be to train models that can be run locally. This would at least solve the first five of the issues mentioned above. To this end, we have experimented with several large language models available on Hugging Face.\footnote{\url{https://huggingface.co/}} A general experience so far is that even the largest of the general LLMs perform poorly on autoformalization tasks and, perhaps surprisingly, large models specifically trained with mathematical content, such as WizardLM-70B\footnote{\url{https://huggingface.co/WizardLM/WizardLM-70B-V1.0}} fare not much better. However, models trained for automated for code generation, such as WizardCoder-Python-34B\footnote{\url{https://huggingface.co/WizardLM/WizardCoder-Python-34B-V1.0}} (apparently the largest local LLM for code generation currently available) show an autoformalization performance comparable to that reported above for text-davinci-003. We plan to systematically develop and evaluate this approach in the near future. 

The ideal would be to create a language model specifically trained on large amounts of data for the task of autoformalization. Work towards such models is done, e.g., in \cite{WBKU}. However, until such pretrained models are available, the approach discussed in this paper seems to offer a workable alternative for certain applications.

\section{Acknowledgements}

We thank our three anonymous referees for several comments that helped in improving the presentation of the paper, along with constructive criticism concerning its content.

\bibliographystyle{eptcs}
\bibliography{diprochegpt}

\newpage

\section{Appendix}

We give here the precise prompt for the assistant used in the autoformalization experiments with GPT-4-Turbo, along with the table listing the results.

We used the following prompt (typesetting adapted for the sake of the reader):

\begin{tcolorbox}
    Given a sentence, translate it into the format [type,subtype,formalization], where ``type'' is one of the following: ``claim'', ``assumption'', ``variable declaration'', ``goal declaration''.

``variable declaration'' has the subtypes ``plain'' and ``with additional assumption'', for statements in which an assumption is made about the introduced variables beyond the type declaration. 

``formalization'' works as follows:

\begin{itemize}
\item For a plain claim, it is just a formalization of the statement in first-order logic. 
\item For an assumption, it is a formalization in first-order logic of the assumed statement.
\item For a plain variable declaration, it is a list of pairs [var,type], where var is the variable to be declared and type is the type assigned to it. 
\item For a variable declaration with additional assumption, it is a pair [declarations,assumption] consisting of the list of declarations as for a plain declaration and a formalization of the assumption. A mere existence claim is not a declaration; thus, "there is an integer x such that ..." is a claim, not a declaration. Declarations are indicated by words such as "Let", "pick", "Choose" etc.
\item If the given sentence contains free variables -- that are not quantified over in the sentence -- they should remain free in the formalization.
\end{itemize}
Return only the triple [type,subtype,formalization], nothing else.

Use $\emptyset$ for the empty set, $c(X)$ for the complement of $X$, $\cup$ for the union of sets, $\cap$ for the intersection of sets, $=$ for equality, $\subseteq$ for the subset relation, $\in$ for the element relation.
\end{tcolorbox}

The results obtained with this prompt for $50$ example sentences can be found in the below table.

\begin{longtable}{|c|p{13em}|l|l|c|}
\hline
Nr & \multicolumn{1}{c|}{Input} & \multicolumn{1}{c|}{Output type} & \multicolumn{1}{c|}{Output formalization} &  +/- \\
\hline
\hline
1 & Let $A$ be a set. & decl/plain & \makecell{[[$A$,set]]} & + \\
\hline
2 & Let $A$ be a non-empty set. & decl/assmpt & \makecell{[[[$A$, set]], $A\neq\emptyset $]} & + \\
\hline
3 & Let $A$, $B$, $C$, be sets. & decl/plain  & \makecell{[[$A$, set], [$B$, set], [$C$, set]]} & + \\
\hline
4 & Let $A$, $B$, $C$ be sets such that  $A\cup B=c(C)$. & decl/assmpt & \makecell{[[[$A$, set], [$B$, set], [$C$, set]], \\ $A\cup B=c(C)$]]} &+ \\
\hline
5 & Let $Q$, $X$ and $U$ as well as $P$ be non-empty sets. & decl/assmpt & \makecell{[[[$Q$, set], [$X$, set], [$U$, set], [$P$, set]], \\ $(Q\neq \emptyset)\wedge(X\neq\emptyset)$\\$\wedge(U\neq\emptyset)\wedge(P\neq\emptyset)$]} & + \\
\hline
6 & Let $G$, $H$ be sets with non-empty intersection & decl/assmpt & \makecell{[[[$G$, set], [$H$,set]], \\ $\exists{x}(x\in G \wedge x\in H)$]} & + \\
\hline 
7& Let $P$, $Q$ be disjoint sets. & decl/assmpt & \makecell{[[[$P$, set], [$Q$, set]], \\ $P\cap Q=\emptyset $]} & + \\
\hline 
8& Let $X$ and $Y$ be sets  and let $U$ be their union. & decl/assmpt & 
\makecell{[[[$X$, set], [$Y$, set], [$U$, set]], \\$U=X\cup Y$]} & + \\
\hline 
9& Let $A$, $B$ be sets with intersection $K$. & decl/assmpt & \makecell{[[[$A$, set], [$B$,set], [$K$, set]], \\$A\cap B=K$]} & + \\
\hline
10& Suppose that $A=B$. & assmpt & $A=B$ & + \\
\hline
11 & Let $x$ be an element of $X$. & decl/assmpt & [[$x$,element],$x\in X$] & + \\
\hline
12&Assume that $A$ and $B$ are disjoint. & assmpt & $A\cap B=\emptyset$ & + \\
\hline
13&Suppose that the union of $A$ and $B$   is equal to their intersection. & assmpt & $A\cup B=A\cap B$ & + \\
\hline
14 &Let us assume that the complement   of $Z$ contains $u$. & assmpt & $u\in c(Z)$ & + \\
\hline
15&Suppose that $m$ is not contained in $M$. & assmpt & $m\in M$ & + \\
\hline
16&Let $A$ be a subset of $B$. & decl/assmpt & [[[$A$,set]],$A\subseteq B$]] & + \\ 
\hline
17&Suppose that $A$ is a subset of $B$. & assmpt & $A\subseteq B$ & + \\
\hline
18&Suppose that $A$ is not a subset of $B$. & assmpt & $\neg(A\subseteq B)$ &+\\
\hline
19&Suppose that neither $A$ is a subset   of $B$ nor vice versa. & assmpt & $\neg(A\subseteq B)\wedge\neg(B\subseteq A)$ & + \\
\hline 
20&Let us take it as given that $x$   belongs to $A$, but not to $B$. & assmpt & $(x\in A)\wedge(x\notin B)$ & + \\
\hline
21& It follows that $A$ and $B$ are equal. & claim & $A=B$ &+ \\
\hline
22&Consequently, $A$ is not equal to $A$. & claim & $A\neq A$ & + \\
\hline 
23&Thus, $A$ is an element of itself. & claim & $A\in A$ & + \\
\hline
24& $A$ is an element of itself and it isn't. & claim & $(A\in A)\wedge\neg(A\in A)$ & + \\
\hline
25&If $A$ is not empty,  then $B$ is a subset of $Q$. & claim & $(A\neq\emptyset)\rightarrow(B\subseteq Q)$ & + \\
\hline 
26&Each element of $A$ is also an element of $O$. & claim & $\forall{x}(x\in A\rightarrow x\in O)$ & + \\
\hline
27&No element of $c(X)$ is   an element of $(A\cup B)\cap C)$ & claim & $\forall{x}(x\in c(X)\rightarrow x\notin (A\cup B)\cap C)$ & + \\
\hline 
28&There is a set $A$ such that $A\cap B=A\cap C$. & claim & $\exists{A}(A\cap B=A\cap C)$ & + \\
\hline
29&There is no set $A$ such that $A\cap B=A\cap C$ & claim & $\neg\exists{A}(A\cap B=A\cap C)$ & + \\
\hline
30&From this, we get that, if $A$ is  not empty, then $B$ is. & claim & $(A\neq\emptyset)\rightarrow(B\neq\emptyset)$ & - \\
\hline
31&From this, we get that, if $A$ is   not empty, then $B$ is empty. & claim & $(A\neq\emptyset)\rightarrow(B=\emptyset)$ & + \\
\hline
32& By reductio, every subset of $A$   is also a subset of $T$. & claim & $\forall{X}(X\subseteq\rightarrow X\subseteq T)$ & + \\
\hline
33 & For all sets $X$, $Y$, $Z$ it holds that: If it holds that $Y$ is a subset of $Z$, then it also holds that the intersection of $X$ and $Y$ is a subset of the intersection of X and Z. & claim & $\forall{X}\forall{Y}\forall{Z}((Y\subseteq Z)\rightarrow(X\cap Y\subseteq X\cap Z))$ & + \\
\hline
34&We will now show that $A$ is in fact empty. & goal & $A=\emptyset$ & + \\
\hline 
%
35&We need to show that $x$ belongs to $K$. & goal & $x\in K$ &  + \\
\hline 
36&We need to demonstrate that,   whenever $A\cap B=A\cup B$, then $A=B$. & goal & $(A\cap B=A\cup B)\rightarrow (A=B)$ & + \\ 
\hline 
37&Our goal is to see that $x$ and $y$ belong to $Q$ & goal & $(x\in Q)\wedge (y\in Q)$ & +  \\
\hline 
38 & To see this, first suppose that $A$ is empty while $B$ is not. & assmpt & $(A=\emptyset)\wedge(B\neq\emptyset)$ & + \\
\hline 
39& Now it remains to establish the   pairwise disjointness of $U$, $V$ and $W$. & goal & \makecell{$(U\cap V=\emptyset)\wedge(U\cap W=\emptyset)$\\$\wedge(V\cap W=\emptyset)$} & + \\
\hline 
40&If $x\in A$, then $x\in (A\cup B)$ & claim & $\forall{x}(x\in A\rightarrow x\in (A\cup B))$ & + \\
\hline 
41& As we will presently show, the complement of $A\cup B$ equals the intersection of the complements of $A$ and $B$. & goal & $c(A\cup B)=c(A)\cap c(B)$ & + \\
\hline 
42&It is thus excluded that $A$ is a subset of $B$. & claim & $\neg(A\subseteq B)$ & + \\
\hline 
43&Therefore, we get that,   if $A$ is empty, then $B$ is not. & claim & $(A=\emptyset)\rightarrow(B\neq\emptyset)$ & + \\
\hline 
44& If $X$ and $Y$ are non-empty and disjoint then both are subsets of $U$. & claim & \makecell{$(X\neq\emptyset\wedge Y\neq\emptyset\wedge X\cap Y=\emptyset)\rightarrow$ \\ $(S\subseteq U\wedge Y\subseteq U)$} & + \\
\hline
45 & It now follows that at least one of   $A$ and $B$ must be empty. & claim & $(A=\emptyset)\vee(B=\emptyset)$ & + \\
\hline
46& From this, we get that exactly one of $A$ and $B$ is empty. & claim &\makecell{$((A=\emptyset)\wedge(B\neq\emptyset))$ \\ $\vee((A\neq\emptyset)\wedge(B=\emptyset))$} & + \\
\hline
47& Let $A$, $B$, $C$ be sets and additionally pick $D$ to be a non-empty set. & decl/assmpt &\makecell{[[[$A$,set], [$B$,set], [$C$, set], [$D$, set]], \\ $D\neq\emptyset$]} & + \\
\hline
48&Consider sets $A$, $B$, $C$ satisfying $A\cap B=c(C)$ & decl/assmpt & \makecell{[[[
$A$,set], [$B$, set], [$C$,set]], \\ $A\cap B=c(C)$]} & + \\
\hline 
49&If one of $A$, $B$ is equal to $C$, then $C\subseteq(A\cup b)$ & claim & \makecell{$((A=C)\vee(B=C))$\\$\rightarrow(C\subseteq(A\cup B))$} & + \\
\hline
50&Thus, the empty set is disjoint from every set.&claim&$\forall{X}(\emptyset\cap X=\emptyset)$ & + \\
\hline
\end{longtable}

\end{document}